\documentclass[sigconf]{acmart}
\AtBeginDocument{%
  }

\copyrightyear{2026}
\acmYear{2026}
\setcopyright{cc}
\setcctype{by}
\acmConference[WSDM '26]{Proceedings of the Nineteenth ACM International Conference on Web Search and Data Mining}{February 22--26, 2026}{Boise, ID, USA}
\acmBooktitle{Proceedings of the Nineteenth ACM International Conference on Web Search and Data Mining (WSDM '26), February 22--26, 2026, Boise, ID, USA}
\acmPrice{}
\acmDOI{10.1145/3773966.3777928}
\acmISBN{979-8-4007-2292-9/2026/02}

\usepackage{algorithm}
\usepackage{algorithmic}
\usepackage{amsmath}
\usepackage{multirow}
\usepackage{booktabs}
\usepackage{lipsum}

\begin{document}

\title{MolEdit: Knowledge Editing for Multimodal Molecule Language Models}

\author{Zhenyu Lei}
\affiliation{%
  \institution{University of Virginia}
  \city{Charlottesville}
    \country{USA}
}
\email{vjd5zr@virginia.edu}

\author{Patrick Soga}
\affiliation{%
  \institution{University of Virginia}
  \city{Charlottesville}
    \country{USA}
}
  \email{zqe3cg@virginia.edu}

\author{Yaochen Zhu}
\affiliation{%
  \institution{University of Virginia}
  \city{Charlottesville}
    \country{USA}
}
  \email{uqp4qh@virginia.edu}

\author{Yinhan He}
\affiliation{%
  \institution{University of Virginia}
  \city{Charlottesville}
    \country{USA}
}
  \email{nee7ne@virginia.edu}

\author{Yushun Dong}
\affiliation{%
  \institution{Florida State University}
  \city{Tallahassee}
    \country{USA}
}
  \email{yushun.dong@fsu.edu}

\author{Jundong Li}
\affiliation{%
  \institution{University of Virginia}
  \city{Charlottesville}
    \country{USA}
}
  \email{jundong@virginia.edu}

\renewcommand{\shortauthors}{Lei et al.}

\begin{abstract}
Understanding and continuously refining multimodal molecular knowledge is crucial for advancing biomedicine, chemistry, and materials science. Molecule language models (MoLMs) have become powerful tools in these domains, integrating structural representations (e.g., SMILES strings, molecular graphs) with rich contextual descriptions (e.g., physicochemical properties, biomedical applications). However, MoLMs can encode and propagate inaccuracies due to outdated web-mined training corpora or malicious manipulation, jeopardizing downstream discovery pipelines. While knowledge editing has been explored for general-domain AI, its application to MoLMs remains uncharted, presenting unique challenges due to the multifaceted and interdependent nature of molecular knowledge. In this paper, we take the first step toward MoLM editing for two critical tasks: molecule-to-caption generation and caption-to-molecule generation. To address molecule-specific challenges, we propose MolEdit, a powerful framework that enables targeted modifications while preserving unrelated molecular knowledge. MolEdit combines a Multi-Expert Knowledge Adapter that routes edits to specialized experts for different molecular facets with an Expertise-Aware Editing Switcher that activates the adapters only when input closely matches the stored edits across all expertise, minimizing interference with unrelated knowledge. To systematically evaluate editing performance, we introduce MEBench, a comprehensive benchmark assessing multiple dimensions, including Reliability (accuracy of the editing), Locality (preservation of irrelevant knowledge), and Generality (robustness to reformed queries).  Across extensive experiments on two popular MoLM backbones, MolEdit delivers up to $18.8 \%$ higher Reliability and $12.0 \%$ better Locality than state-of-the-art editing baselines while maintaining efficiency. Our findings chart a clear path toward safer, continuously updatable scientific foundation models.  The code is available at: https://github.com/LzyFischer/MolEdit.

\end{abstract}

\ccsdesc[500]{Computing methodologies~Machine learning}

\keywords{Molecule Language Model, Multimodal, Knowledge Editing}


\maketitle

\section{Introduction}
\begin{figure}[t!]
    \vspace{5pt}
    \centering
    \includegraphics[width=\columnwidth]{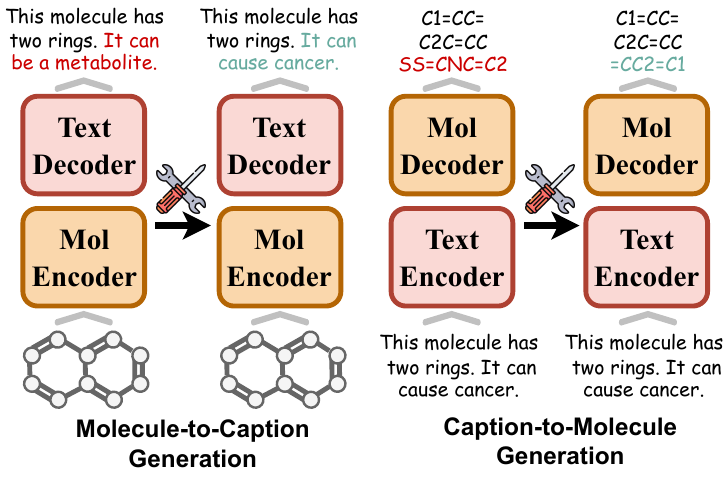}
    \caption{An illustration of MoLM editing for two tasks: correcting inaccurate captions in \textit{molecule-to-caption generation} and fixing mismatched or invalid molecules in \textit{caption-to-molecule generation}.}
    \label{fig:teasor}
\end{figure}

Understanding and continuously refining molecular knowledge is crucial across various scientific fields, such as biomedicine~\cite{zhang2024scientific, pei2024leveraging}, chemistry~\cite{liao2024words, xiao2024bridging}, and materials science~\cite{lei2024materials}. Pre-trained and fine-tuned on diverse multimodal data sourced from web such as PubChem~\cite{edwards2024l+}, molecule language models (MoLMs) encode multimodal molecular knowledge, encompassing structural representations (e.g., SMILES strings, molecular graphs) and rich contextual descriptions (e.g., physicochemical properties, biomedical applications)~\cite{su2022molecular, pei20243d, cao2023instructmol}.
With rich knowledge integrated, MoLM encoders ingest heterogeneous inputs while decoders generate outputs across modalities, enabling key applications such as molecule-to-caption generation and caption-to-molecule design~\cite{luo2023molfm}.
However, during knowledge integration, inaccurate or misleading information can be introduced from outdated web-mined training corpora~\cite{deng2024chemical}, posing risks in downstream applications. For instance, 
Tox21~\cite{tice2013improving} initially labeled several toxic molecules as non-toxic based on early assays, which MoLMs may memorize and propagate, potentially leading to critical errors in drug discovery pipelines.
Hence, there is a pressing need to refine MoLMs to correct inaccurate or misleading knowledge from outdated training data or malicious manipulation, as shown in Figure~\ref{fig:teasor}.
Recently, knowledge editing has emerged as an efficient approach to modifying specific knowledge while preserving other information~\cite{wang2024knowledge, zhang2024comprehensive, mazzia2024survey}. It has been widely applied to general-domain large language models~\cite{de2021editing, meng2022mass} and multimodal language models~\cite{huang2024vlkeb, chengetal2023edit}.

However, the adaptation of knowledge editing to MoLMs remains largely unexplored, which presents two inherent molecule-specific challenges.
First, molecular knowledge is inherently \emph{multifaceted}~\cite{cao2023instructmol}. A single compound consists of multiple functional groups such as aromatic rings, heteroatoms, and chiral centers, each carrying its own reactivity and toxicity signals, while its caption comprises distinct descriptive components such as functional descriptions (e.g., "antiviral diterpenoid") and provenance notes (e.g., "isolated from Taxus brevifolia"). Each of these elements represents a specific aspect of molecular properties and may exhibit varying sensitivities to editing. Consequently, modifying multifaceted molecular knowledge poses the risk of over-editing certain aspects while under-editing others~\cite{javadi2024knowledge, zheng2023can}.
Second, shared functional groups and contextual descriptions create dense \emph{interdependencies} among molecules, so editing one molecule’s knowledge can unintentionally affect others with similar features. This makes it difficult to ensure edits remain localized, violating the principle of locality, where modifications should only impact the intended target. For instance, both celecoxib (an anti-inflammatory) and sulfamethoxazole (an antibiotic) share a sulfonamide group, yet they differ markedly in toxicity and side-effect profiles. Editing the model's knowledge of celecoxib might as well alter the model’s knowledge of sulfamethoxazole due to their structural similarity, even though the latter should not be edited.


To address these challenges, we propose MolEdit, a simple yet powerful framework for editing multimodal MoLMs\footnote{We focus on multimodal MoLMs because they are more challenging and requires edits across both structural and textual representations of molecules.} in caption generation and molecule design tasks. Our approach enables precise, targeted updates to compositional molecular knowledge while preserving unrelated information.
%
To address the first challenge, we design a \textit{Multi-Expert Knowledge Adapter} (MEKA) that decomposes a molecule or caption into distinct facets and directs different facets of molecular knowledge to specialized editing experts within a lightweight Mixture-of-Expert layer, thus enabling fine-grained control over multifaceted updates. For the second challenge, we introduce an \textit{Expertise-Aware Editing Switcher} (EAES), which maintains a memory bank of edited molecular knowledge and activates the knowledge adapter only when the input closely matches the stored edits across all expertise, thus minimizing interference with unrelated knowledge. Furthermore, since incorrect outputs can arise from interactions between different modalities, MolEdit edits both encoders and decoders to ensure comprehensive refinement.

To enable systematic evaluation, we introduce MEBench, the first benchmark for editing MoLMs, which rigorously assesses reliability (editing accuracy), locality (preservation of unrelated knowledge), and generality (consistency across varied textual descriptions). Extensive experiments on two popular MoLM backbones show that MolEdit surpasses fine-tuning, gradient-based, and memory-based baselines across all three MEBench dimensions. Overall, our key contributions are as follows:
\begin{itemize}
    \item \textbf{Problem Formulation}: We present the first systematic exploration of \emph{knowledge editing} in molecule language models, clearly delineating two molecule-specific knowledge editing challenges including multifaceted expertise and cross-molecule interdependence.

    \item \textbf{Benchmark Construction}: We release \emph{MEBench}, a rigorously curated dataset that quantifies editing success along three dimensions including \emph{reliability}, \emph{locality}, and \emph{generality}, covering both molecule-to-caption and caption-to-molecule tasks.

    \item \textbf{Framework Design}: We introduce \emph{MolEdit}, a powerful framework that combines a \emph{Multi-Expert Knowledge Adapter} for facet-level editing with an \emph{Expertise-Aware Editing Switcher} that gates updates based on fine-grained similarity, thereby enforcing fine-grained control and locality.
    
    \item \textbf{Experimental Evaluation}: Through extensive studies on two strong MoLM backbones, MolEdit delivers consistent gains over state-of-the-art gradient-based and external-memory-based baselines on every MEBench metric, demonstrating its effectiveness.
\end{itemize}


\section{Related Work}
\noindent\textbf{Molecule Language Models.} Building on the success of general-domain language models, \emph{molecule language models} (MoLMs) aim to learn unified representations that cover structure, property and function~\cite{liu2024scientific, pei2024leveraging}. Early single-modal molecule language models such as ChemBERTa and MolT5 concentrated solely on SMILES strings, whereas more recent work shifted into \textit{multimodality} including graph and SMILES encoders that capture molecule structural knowledge and text encoders ingest synthesis notes or pharmacological annotations~\cite{liu2023multi, pei2023biot5}. To encode multimodal molecular knowledge, two broad pre-training paradigms have emerged.  \emph{Generative} schemes treat every modality as a sequence and train an autoregressive decoder that can translate between them, enabling direct sequence-to-sequence tasks~\cite{fang2023mol, zeng2022deep, christofidellis2023unifying, zhao2023gimlet}.  By contrast, \emph{contrastive} schemes retain separate encoders and maximize cross-modal agreement, yielding greater architectural flexibility and lower latency at inference~\cite{su2022molecular, luo2023molfm, liu2023molca, li2024towards, liu2024git, luo2024learning}. After pre-training, MoLMs are fine-tuned for downstream tasks such as molecule-to-caption generation, caption-to-molecule design, virtual screening, and property prediction~\cite{su2022molecular, li2024empowering, gong2024text, liu2024drugllm, soga2025deep}. However, both the diverse pre-training corpora and task-specific fine-tuning stages can inject stale or spurious facts such as mis-annotated toxicity labels or incorrect stereochemistry~\cite{dong2022calibrating, huang2024can}. These issues underscore the need for techniques that can edit a model’s knowledge without expensive retraining.

\noindent\textbf{Knowledge Editing.} A naive way to correct model errors is to \emph{fine-tune} on a small patch dataset, while it can cause costly retraining and catastrophic forgetting of unrelated facts. Knowledge editing therefore focuses on making targeted, data-efficient corrections to a model’s internal parameters while avoiding collateral damage to unrelated facts~\cite{mazzia2024survey, zhang2025resolving}. There are several types of knowledge editing methods. \emph{Gradient-based} editors adjust weights directly via one or a few back-prop steps, either with meta-learned optimizers~\cite{sinitsin2020editable, cheng2024editing, mitchell2021fast} or with locate-then-edit strategies that first identify influential sub-spaces~\cite{meng2022locating, meng2022mass, ni2023forgetting}. Although effective, they might still distort neighbouring knowledge. \emph{External-memory} approaches instead graft new information onto adapters, side networks, or retrieval memories—reducing interference at the cost of extra parameters~\cite{mitchell2022memory, hartvigsen2024aging, huang2023transformer, wang2024lemoe, zheng2023can, madaan2022memory}.  Recent work has begun extending these ideas to multimodal frameworks~\cite{zeng2024visual, cheng2023can}, but none of them are designed to address the \emph{multifaceted} and highly \emph{interdependent} challenge of multimodal chemical knowledge. MolEdit fills this gap with MEKA which routes distinct facets to specialised experts and EAES that activates those experts only when all facets of the query align with a stored edit, thus achieving fine-grained updates while preserving surrounding chemistry.


\section{Preliminary}
\paragraph{\textbf{Notations.}} Let $\mathcal{G} = (\mathcal{V}, \mathcal{E})$ represent a molecular graph, where $\mathcal{V}$ denotes the set of atoms (nodes) and $\mathcal{E}$ represents covalent bonds (edges). Each molecular graph consists of $N_g$ subgraphs ${g_i}_{i=1}^{N_g}$, each corresponding to a functional group. The molecular structure can also be expressed as a SMILES string, denoted as $\mathcal{S}$. Additionally, each molecule is associated with a caption $\mathcal{T} = {t^1, ..., t^{N_t}}$ containing $N_t$ textual descriptions.

\paragraph{\textbf{Molecule Language Model.}} Given molecular representations $\mathcal{G}$, $\mathcal{S}$, and captions $\mathcal{T}$, molecule language models learn aligned cross-modal representations by utilizing a molecule and a text encoder during pretraining. A task-specific decoder is then appended for downstream generation tasks.

\paragraph{\textbf{Caption Generation.}} For molecule-to-caption generation task, a pretrained text decoder is appended to the molecule encoder and fine-tuned to generate captions $\mathcal{T}$ given a molecular representation $\mathcal{G}$, $\mathcal{S}$. We denote the fine-tuned MoLMs for this task as $f_{cap}$.

\paragraph{\textbf{Molecule Generation.}} Similarly, for the caption-to-molecule generation task, a pretrained molecule generation decoder is appended to the text encoder and fine-tuned to generate SMILES representations $\mathcal{S}$ given a caption $\mathcal{T}$~\footnote{We follow the definition of existing MoLMs~\cite{luo2023molfm, luo2024learning, edwards2022translation}}. We denote the fine-tuned MoLMs for this task as $f_{gen}$.

\section{Editing Molecule Language Model} 
In this section, we first introduce the task of editing MoLMs for molecule and caption generation (\S\ref{sec:task}). We then present MEBench, the first benchmark for evaluating MoLM editing (\S\ref{sec:dataset}), followed by MolEdit, a simple yet powerful framework designed to address molecule-specific challenges (\S\ref{sec:model}).

\subsection{Task Definition}
\label{sec:task}

\paragraph{\textbf{Editing Caption Generation.}} 
MoLMs may produce inaccurate outdated captions that require correction. To assess editing effectiveness, we evaluate caption generation along two dimensions. First, \textbf{Reliability} measures how well the edited captions align with the expert-curated ground truth. Given a dataset $\mathcal{D}^{\mathcal{T}}{\text{edit}}$ containing molecules with initially incorrect captions, we compute the semantic similarity between the captions generated by the edited MoLMs, $\tilde{f}_{cap}$, and the target captions $\mathcal{T}$:
\begin{equation}
\mathcal{M}^{\mathcal{T}}_{rel}=\mathbb{E}_{\left(\mathcal{G},\mathcal{S},\mathcal{T}\right) \sim \mathcal{D}^{\mathcal{T}}_{\text{edit}}} (\textbf{SIM}_T(\tilde{f}_{cap}(\mathcal{G}, \mathcal{S}), \mathcal{T})),
\end{equation}
where $\textbf{SIM}_T$ is the metric to measure text similarity. Secondly, \textbf{Locality} preserves existing knowledge by minimizing the deviations in unedited captions. For a dataset $\mathcal{D}^{\mathcal{T}}_{\text{loc}}$ with knowledge unrelated to $\mathcal{D}^{\mathcal{T}}_{\text{edit}}$, we compare outputs before and after editing:
\begin{equation}
\mathcal{M}^{\mathcal{T}}_{loc}=\mathbb{E}_{ \left(\mathcal{G},\mathcal{S}\right) \sim \mathcal{D}^{\mathcal{T}}_{\text{loc}}} (\textbf{SIM}_T(\tilde{f}_{cap}(\mathcal{G, S}), f_{cap}(\mathcal{G, S}))).
\end{equation}
\paragraph{\textbf{Editing Molecule Generation.}}
MoLMs can also generate invalid molecule SMILES that require correction. Similar to editing caption generation, we evaluate the \textbf{Reliability} and \textbf{Locality} with edited MoLMs for molecule generation $\tilde{f}_{gen}$:
\begin{equation}
\mathcal{M}^{\mathcal{S}}_{rel}=\mathbb{E}_{\left(\mathcal{S},\mathcal{T}\right) \sim \mathcal{D}^{\mathcal{S}}_{\text{edit}}} (\textbf{SIM}_G(\tilde{f}_{gen}(\mathcal{T}), \mathcal{S})),
\end{equation}
\begin{equation}
\mathcal{M}^{\mathcal{S}}_{loc}=\mathbb{E}_{ \left(\mathcal{T}\right) \sim \mathcal{D}^{\mathcal{S}}_{\text{loc}}} (\textbf{SIM}_G(\tilde{f}_{gen }(\mathcal{T}), f_{gen}(\mathcal{T}))),
\end{equation}
where $\mathcal{D}^{\mathcal{S}}_{\text{edit}}$ contain molecules requiring knowledge editing and $\mathcal{D}^{\mathcal{S}}_{\text{loc}}$ ought to remain unchanged during editing. $\textbf{SIM}_G$ is the metric to measure molecule similarity. Additionally, to prevent the model from merely memorizing descriptions during training, semantically equivalent descriptions should be designed to generate the same molecule. To evaluate this, we assess the model's output consistency using a \textbf{generality} dataset, as shown below:
\begin{equation}
\mathcal{M}^{\mathcal{S}}_{gen}=\mathbb{E}_{\substack{
(\mathcal{T}_r) \sim \mathcal{N}(\mathcal{T}) \\  (\mathcal{S,T}) \sim \mathcal{D}^{\mathcal{S}}_{\text{edit}}}} (\textbf{SIM}_G(\tilde{f}_{gen}(\mathcal{T}_r), \mathcal{S})),
\end{equation}
where the $\mathcal{N}(\cdot)$ denotes the description generalization set.

\begin{figure}
    \centering
    \includegraphics[width=\linewidth]{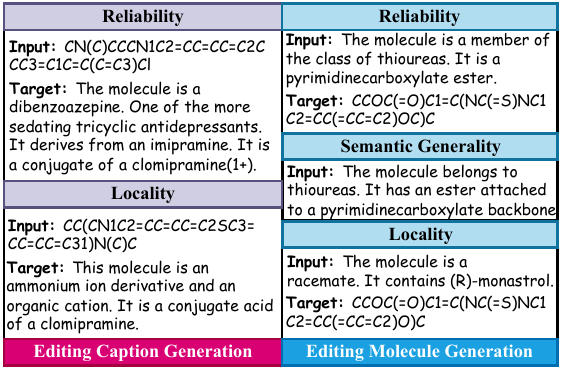}
    \caption{A sample illustration of MEBench. It includes three evaluation dimensions for two tasks: Reliability (molecules requiring editing), Locality (similar but untargeted molecules), and Generality (semantically equivalent but different captions).}
    \vspace{-10pt}
    \label{fig:dataset}
\end{figure}

\begin{figure*}[h]
    \centering
    \includegraphics[width=0.9\textwidth]{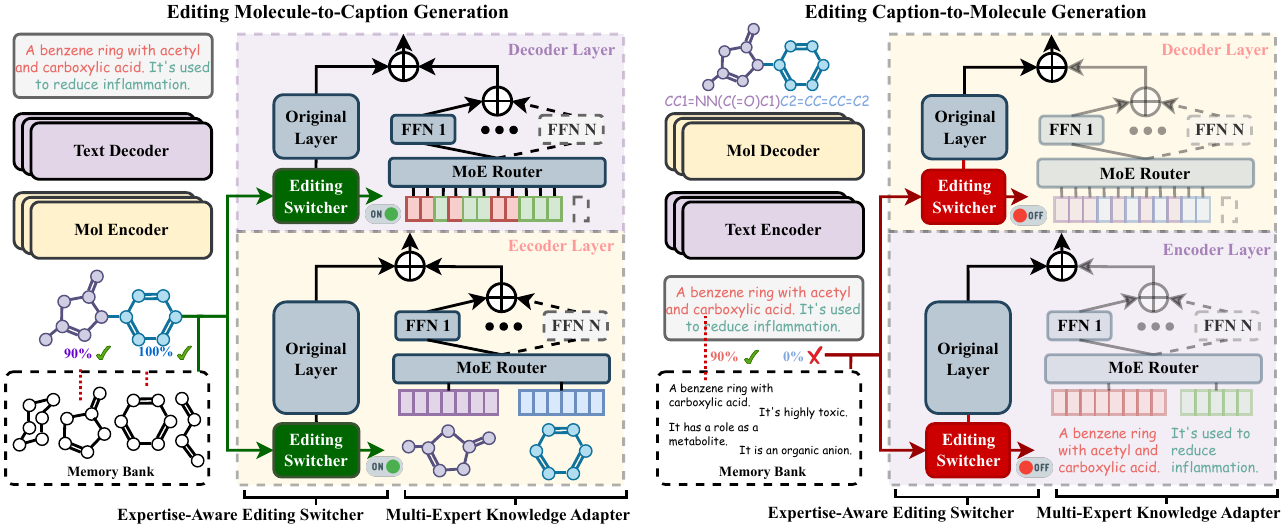}
    \caption{An overview of MolEdit to edit MoLMs for molecule/caption generation by modifying a chosen layer in either the encoder and decoder. It is composed of two components: (1) Multi-Expert Knowledge Adapter (MEKA) and (2) Expertise-Aware Editing Switcher (EAES). Specifically, MEKA utilizes expertise-wise MoE for encoder and token-wise MoE for decoder to route expertise to different editing experts (instantiated as FFN). EAES stores edited knowledge expertise (functional groups/descriptions) and activates MEKA only when all input expertise finds a similar match in its memory bank during inference, thereby preserving unrelated knowledge.
}
    \label{fig:overview}
\end{figure*}

\subsection{Benchmark Construction}
\label{sec:dataset}
This subsection provides a brief overview of the MEBench construction procedure. For clarity, Figure \ref{fig:dataset} depicts a representative sample drawn from the final benchmark.

\subsubsection{Editing Caption Generation.}
\paragraph{\textbf{Reliability.}}
To assess the effectiveness of knowledge editing, we construct a caption reliability dataset, $\mathcal{D}^{\mathcal{T}}_{\text{edit}}$. 
We first identify suboptimal entries (BLEU-2 score~\cite{wieting2019beyond} $< 0.2$) shared across multiple MoLMs within the widely used molecule dataset (CheBI-20~\cite{edwards2022translation}) for caption generation, using its ground truth captions as the editing targets. Additionally, to enable a more fine-grained evaluation of editing capabilities (as discussed in Section~\ref{method_valid}), the targets are decomposed into distinct descriptions, each capturing a specific aspect of molecular knowledge.

\paragraph{\textbf{Locality.}}
To assess the ability of MoLMs to preserve existing knowledge, we construct a locality dataset, $\mathcal{D}^{\mathcal{T}}_{\text{loc}}$. Specifically, we extract high-accuracy entries (BLEU-2 score $> 0.95$) from the training set to serve as the basis for this dataset. Since knowledge editing is more likely to affect knowledge similar to the editing target, we select locality samples with high similarity to those in the reliability dataset, making the editing task more challenging. Following the design of UnKEBench~\cite{deng2024unke}, both the reliability and locality datasets contain unstructured knowledge, as molecule captions are complex and involve multiple entities.

\subsubsection{Editing Molecule Generation.}
\paragraph{\textbf{Reliability.}}
To assess the effectiveness of knowledge editing in improving molecule SMILES generation, we construct the molecule reliability dataset, $\mathcal{D}^{\mathcal{S}}_{\text{edit}}$. Following a similar approach to caption editing dataset construction, we identify suboptimal entries (MACCS FTS~\cite{zhang2024atomas} $< 0.2$) shared across multiple MoLMs within the same dataset. The ground truth SMILES representations in the dataset serve as the editing targets.

\paragraph{\textbf{Locality.}}
To construct the generation locality dataset, $\mathcal{D}^{\mathcal{S}}_{\text{loc}}$, we select high-accuracy entries (MACCS FTS $> 0.95$) from multiple MoLMs within the training set. To ensure a rigorous evaluation of the model’s ability to preserve existing knowledge, we choose entries with the highest similarity to those in the reliability dataset.

\paragraph{\textbf{Generality.}} To assess a model's ability to generalize across different phrasings, we construct the generality dataset, ${\mathcal{N(\mathcal{T})}}$. This dataset consists of rephrased captions from the molecule reliability dataset, evaluating whether semantically equivalent descriptions consistently generate the same molecular structures.

\subsection{Methodology}
\label{sec:model}

To tackle the challenges of editing MoLMs, we propose MolEdit, a simple yet powerful framework designed to enable localized updates to multifaceted molecular knowledge. As shown in Figure~\ref{fig:overview}, MolEdit comprises two key components: (1) \textit{Multi-Expert Knowledge Adapter} - This module disentangles and customizes the editing of diverse molecular knowledge (e.g., functional groups in molecules, descriptive elements in captions) by dynamically routing them to specialized editing experts via a Mixture-of-Experts (MoE) architecture. (2)\textit{Expertise-Aware Editing Switcher} -  This component ensures edits are applied only to highly relevant inputs by leveraging a memory bank of expertise embeddings, activating modifications only when the input exhibits substantial overlap with stored edits.

\subsubsection{Multi-Expert Knowledge Adapter.}
Since errors can originate from both input and output modalities~\cite{chengetal2023edit}, MolEdit enables knowledge editing by wrapping selected layers in both the encoder and decoder with adapters, allowing localized parameter updates. Taking the molecule generation editing task as an example, when editing the encoder at layer $l$, we employ $P$ distinct experts to perform customized edits for different descriptions in the input. Each expert is instantiated as a feed-forward network (FFN) layer. To achieve this, a gating function dynamically routes the $(l-1)$-th layer embeddings of each description to the appropriate expert, illustrated as,
\begin{equation}
\mathcal{P}^n=\operatorname{top}_k\left(\operatorname{softmax}\left(\mathbf{W}_g \cdot \Sigma_{i \in t_n} (z_i^{l-1}) / |t^n|+\epsilon\right)\right),
\end{equation}
where $t^n$ is the $n$-th description and $z_i^{l-1}$ is the embedding of $i$-th token in $t^n$ at $(l-1)$-th layer. $\mathbf{W}_g$ is the trainable weights in gate decision, while $\epsilon$ denotes the noise term. The top$_k(\cdot)$ operator zeros out all but the top-$k$ values. After getting the gate decision vector $\mathcal{G}^n$ of the $n$-th description, the corresponding output is generated through a weighted aggregation of each expert’s computation on $z_i^l$, demonstrated as,
\begin{equation}
z^{l}_i = f^l(z^{l-1}_i) + \sum_{p=1}^P \mathcal{P}^n_p \cdot \mathbf{W}_p \cdot z^{l-1}_i ,\quad i \in t^n,
\end{equation}
where $\mathbf{W}_p$ is the trainable weights. When editing the decoder at layer $l'$, the ground truth expertise segmentation is unavailable. As a result, the MoE adapter is applied to each token, which is expected to route tokens associated with different expertise to the appropriate experts, demonstrated as,
\begin{equation}
\begin{aligned}
\mathcal{P}^i &= \operatorname{top}_k\left(\operatorname{softmax}\left(\mathbf{W}^{'}_g \cdot z_i^{l'-1} +\epsilon\right)\right), \\
z^{l'}_i &= f^{l'}(z^{l'-1}_i)+\lambda \sum_{p=1}^P \mathcal{P}^i_p \cdot \mathbf{W}^{'}_p \cdot z^{l'-1}_i.
\end{aligned}
\end{equation}

Similarly, for caption generation, the input molecule is composed of multiple functional groups, each representing a distinct molecule expertise. We route by functional groups during encoder editing and by token during decoder editing due to the lack of predefined expertise in decoder.


\subsubsection{Expertise-Aware Editing Switcher.}

In order to minimize unintended interference with untargeted molecules that share a few functional groups or descriptions, we design a expertise-aware switching mechanism that only allows activation of the knowledge adapters for molecules with high expertise overlap with edited molecules. Specifically, we keep an expertise-based memory bank that stores the expertise-wise knowledge through the form of encoder embeddings of the edit queries:
\begin{equation}
\begin{aligned}
    \mathcal{Z} = \{\Bar{z}_{n_j}\}, \quad \Bar{z}_{n_j} = \Sigma_i^{i \in t_{n_j}} (z_i^{enc}) / |t_{n_j}|, \\
\end{aligned}
\end{equation}
where $z_i^{enc}$ represents the encoder embedding of the $i$-th token (node) within the $n_j$-th description (functional group) of the $j$-th sample. 
During inference, the switcher compares each expertise in the input to the expertise of stored edits in the memory bank $\mathcal{Z}$. The adapter is activated only when all expertise distances are below a threshold $\epsilon$, ensuring that unrelated molecules remain unaffected:
\begin{equation}
z^l_{i} = \begin{cases}\operatorname{MolEdit}\left(z^{l-1}_i\right) & \text { if } \max_{n}\left(d\left(\Bar{z}_{n}, \mathcal{Z}\right)\right)<\epsilon, \\ f^l\left(z^{l-1}_i\right) & \text {otherwise},\end{cases}
\end{equation}
where $d(\cdot)$ is a distance function.

\begingroup
\renewcommand{\arraystretch}{1.02}
\begin{table*}[t!]
    \centering
    \resizebox{0.9\textwidth}{!}{%
    \begin{tabular}{llcccccc}
        \toprule
        \multirow{2}{*}{} & \multirow{2}{*}{} & \multicolumn{3}{c}{\textbf{MoMu}} & \multicolumn{3}{c}{\textbf{MolFM}} \\
        \cmidrule(lr){3-5} \cmidrule(lr){6-8}
         &  & BLEU-4$\uparrow$ & LEV$\downarrow$ & MACCS$\uparrow$ & BLEU-4$\uparrow$ & LEV$\downarrow$ & MACCS$\uparrow$\\
        \midrule
        \multirow{6}{*}{\textbf{Rel}} 
        & FT (Encoder) & 0.758~\footnotesize{$(\pm 0.022)$}	& 22.974~\footnotesize{$(\pm 3.224)$}	& 0.987~\footnotesize{$(\pm 0.018)$}
& \textbf{0.983}~\footnotesize{$(\pm 0.000)$}	& \textbf{2.015}~\footnotesize{$(\pm 0.018)$}	& \underline{0.998}~\footnotesize{$(\pm 0.001)$} \\
        & FT (Decoder) & 0.711~\footnotesize{$(\pm 0.002)$}	& 29.936~\footnotesize{$(\pm 0.112)$}	& 0.938~\footnotesize{$(\pm 0.002)$}
& 0.894~\footnotesize{$(\pm 0.035)$}	& 11.560~\footnotesize{$(\pm 4.244)$}	& 0.977~\footnotesize{$(\pm 0.013)$} \\
        & FT (All) & 0.781~\footnotesize{$(\pm 0.000)$}	& \underline{19.940}~\footnotesize{$(\pm 0.035)$}	& \textbf{1.000}~\footnotesize{$(\pm 0.000)$}
& \underline{0.977}~\footnotesize{$(\pm 0.011)$}	& \underline{2.554}~\footnotesize{$(\pm 1.091)$}	& 0.995~\footnotesize{$(\pm 0.001)$} \\
        \cmidrule{2-8}
        & MEND & \underline{0.802}~\footnotesize{$(\pm 0.019)$}	& 21.079~\footnotesize{$(\pm 1.360)$}	& 0.834~\footnotesize{$(\pm 0.011)$}
& 0.789~\footnotesize{$(\pm 0.003)$}	& 23.547~\footnotesize{$(\pm 0.125)$}	& 0.869~\footnotesize{$(\pm 0.003)$} \\
         & GRACE & 0.718~\footnotesize{$(\pm 0.000)$}	& 28.331~\footnotesize{$(\pm 0.025)$}	& 0.938~\footnotesize{$(\pm 0.000)$}
& 0.770~\footnotesize{$(\pm 0.001)$}	& 11.464~\footnotesize{$(\pm 15.122)$}	& 0.987~\footnotesize{$(\pm 0.004)$} \\
        & \textbf{MolEdit} & \textbf{0.953}~\footnotesize{$(\pm 0.025)$}	& \textbf{4.667}~\footnotesize{$(\pm 2.154)$}	&\underline{0.989}~\footnotesize{$(\pm 0.008)$}
& 0.975~\footnotesize{$(\pm 0.003)$}	& 2.862~\footnotesize{$(\pm 0.479)$}	& \textbf{1.000}~\footnotesize{$(\pm 0.000)$} \\
        \midrule
        \multirow{6}{*}{\textbf{Loc}}
        & FT (Encoder) & 0.829~\footnotesize{$(\pm 0.001)$}	& 18.786~\footnotesize{$(\pm 0.089)$}	& 0.881~\footnotesize{$(\pm 0.008)$}
& 0.829~\footnotesize{$(\pm 0.001)$}	& 19.622~\footnotesize{$(\pm 0.154)$}	& \underline{0.943}~\footnotesize{$(\pm 0.005)$} \\
        & FT (Decoder) & 0.881~\footnotesize{$(\pm 0.001)$}	& 12.562~\footnotesize{$(\pm 0.094)$}	& 0.936~\footnotesize{$(\pm 0.003)$}
& 0.725~\footnotesize{$(\pm 0.001)$}	& 31.195~\footnotesize{$(\pm 0.203)$}	& 0.826~\footnotesize{$(\pm 0.002)$} \\
        & FT (All) & 0.891~\footnotesize{$(\pm 0.004)$}	& 11.756~\footnotesize{$(\pm 0.413)$}	& \underline{0.949}~\footnotesize{$(\pm 0.005)$}
& 0.803~\footnotesize{$(\pm 0.009)$}	& 21.985~\footnotesize{$(\pm 1.189)$}	& 0.909~\footnotesize{$(\pm 0.006)$} \\
        \cmidrule{2-8}
        & MEND & \underline{0.912}~\footnotesize{$(\pm 0.028)$}	& \underline{9.512}~\footnotesize{$(\pm 3.291)$}	& 0.940~\footnotesize{$(\pm 0.030)$}
& \underline{0.833}~\footnotesize{$(\pm 0.020)$}	& \underline{17.581}~\footnotesize{$(\pm 1.750)$}	& 0.937~\footnotesize{$(\pm 0.014)$} \\
         & GRACE & 0.859~\footnotesize{$(\pm 0.000)$}	& 15.732~\footnotesize{$(\pm 0.005)$}	& 0.937~\footnotesize{$(\pm 0.000)$}
& 0.757~\footnotesize{$(\pm 0.001)$}	& 30.112~\footnotesize{$(\pm 0.121)$}	& 0.894~\footnotesize{$(\pm 0.015)$}\\
        & \textbf{MolEdit} & \textbf{0.980}~\footnotesize{$(\pm 0.007)$}	& \textbf{1.853}~\footnotesize{$(\pm 0.727)$}	& \textbf{0.997}~\footnotesize{$(\pm 0.000)$}
& \textbf{0.918}~\footnotesize{$(\pm 0.034)$}	& \textbf{9.167}~\footnotesize{$(\pm 3.963)$}	& \textbf{0.991}~\footnotesize{$(\pm 0.007)$} \\
        \midrule
        \multirow{6}{*}{\textbf{Gen}}
        & FT (Encoder) & 0.526~\footnotesize{$(\pm 0.013)$}	& 55.008~\footnotesize{$(\pm 0.359)$}	& 0.629~\footnotesize{$(\pm 0.012)$}
& 0.727~\footnotesize{$(\pm 0.021)$}	& \underline{30.209}~\footnotesize{$(\pm 1.162)$}	& 0.697~\footnotesize{$(\pm 0.051)$} \\
        & FT (Decoder) & 0.590~\footnotesize{$(\pm 0.011)$}	& 48.795~\footnotesize{$(\pm 1.465)$}	& 0.753~\footnotesize{$(\pm 0.021)$}
& 0.665~\footnotesize{$(\pm 0.002)$}	& 38.035~\footnotesize{$(\pm 0.172)$}	& 0.649~\footnotesize{$(\pm 0.023)$} \\
        & FT (All) & 0.586~\footnotesize{$(\pm 0.006)$}	& 47.872~\footnotesize{$(\pm 1.192)$}	& 0.745~\footnotesize{$(\pm 0.013)$}
& 0.713~\footnotesize{$(\pm 0.002)$}	& 30.755~\footnotesize{$(\pm 2.094)$}	& 0.691~\footnotesize{$(\pm 0.017)$} \\
        \cmidrule{2-8}
        & MEND & \underline{0.707}~\footnotesize{$(\pm 0.025)$}	& \underline{30.551}~\footnotesize{$(\pm 2.379)$}	& 0.721~\footnotesize{$(\pm 0.009)$}
& \underline{0.731}~\footnotesize{$(\pm 0.000)$}	& 30.945~\footnotesize{$(\pm 0.513)$}	& 0.678~\footnotesize{$(\pm 0.052)$} \\
         & GRACE & 0.703~\footnotesize{$(\pm 0.000)$}	& 32.574~\footnotesize{$(\pm 0.000)$}	& \underline{0.913}~\footnotesize{$(\pm 0.000)$}
& 0.645~\footnotesize{$(\pm 0.000)$}	& 44.521~\footnotesize{$(\pm 0.732)$}	& \underline{0.892}~\footnotesize{$(\pm 0.029)$} \\
        & \textbf{MolEdit} & \textbf{0.842}~\footnotesize{$(\pm 0.028)$}	& \textbf{15.609}~\footnotesize{$(\pm 2.304)$}	& \textbf{0.917}~\footnotesize{$(\pm 0.020)$}
& \textbf{0.796}~\footnotesize{$(\pm 0.020)$}	& \textbf{23.314}~\footnotesize{$(\pm 1.576)$}	& \textbf{0.895}~\footnotesize{$(\pm 0.044)$} \\
        \bottomrule
    \end{tabular}
    }
    \caption{Main results on MEBench for editing MoMu and MolFM in molecule generation, evaluated across three dimensions: Reliability (Rel), Locality (Loc), and Generality (Gen). Each dimension uses three metrics: BLEU-4, LEV, and MACSS. The best and second-best results are shown in \textbf{bold} and \underline{underlined}, respectively. FT denotes fine-tuning.}
    \vspace{-10pt}
    \label{tab:main_smiles}
\end{table*}
\endgroup

\section{Experiments}
We aim to answer three key research questions: \textbf{RQ1}: How does MolEdit perform compared to baselines on MEBench? \textbf{RQ2}: How does each component contribute to MolEdit's performance? \textbf{RQ3}: How can we explain the effectiveness of each module? In addition, we also include qualitative generation examples contrasting MolEdit with baseline outputs. The analysis is presented below.


\subsection{Experiment Settings}
Below we introduce the experiment settings, including MoLM backbones, baselines, metrics, and implementation details.

\subsubsection{MoLM Backbone.}
In this paper, we adopt two widely utilized multimodal MoLMs with different paradigms as the editing backbones: MoMu~\cite{su2022molecular} and MolFM~\cite{luo2023molfm}.

\noindent\textbf{MoMu.} 
MoMu is a multimodal MoLM aligning molecule graphs and text through contrastive learning~\cite{li2021supervision}. MoMu utilizes a GIN encoder for graphs~\cite{xu2018powerful} and a Bert encoder for text~\cite{devlin2018bert}. For generation, MoMu attaches a MolT5 decoder~\cite{edwards2022translation}, enabling both molecule-to-caption and caption-to-molecule tasks. 

\noindent\textbf{MolFM.}
MolFM integrates molecular structures, biomedical text, and knowledge graphs using cross-modal attention to fuse signals across modalities. It uses GraphMVP as the graph encoder~\cite{liu2021pre} and BERT for text, with MolT5 as a unified decoder for caption and molecule generation. Compared with MoMu’s contrastive alignment, MolFM’s attention-based fusion yields richer token-level interactions during pretraining, offering a complementary testbed for molecular knowledge editing.

\subsubsection{Baselines.}
\noindent\textbf{Fine-tune.} Fine-tune adapts pre-trained language models to specific tasks and remains the default knowledge editing baseline~\cite{li2023unveiling, zhong2023mquake, ma2025comprehendedit}. We experiment with three fine-tuning variants: (1) \emph{encoder-only}, (2) \textit{decoder-only}, and (3) \textit{encoder+decoder} editing.

\noindent\textbf{MEND.} MEND~\cite{mitchell2021fast} is an \textit{optimization-based method }that adopts a mechanism referred to as gradient decomposition. It employs auxiliary editing networks that transform model gradients via low-rank decomposition. We adapt MEND to our multimodal setting by applying edits at both encoder and decoder layers.

\noindent\textbf{GRACE.} GRACE~\cite{hartvigsen2024aging} is a \textit{memory-based method} that employs a deferral mechanism to decide whether to activate an adapter by comparing a given input against a codebook of stored edits. Similar to MEND, we extend GRACE to the multi-modal setting.

\subsubsection{Metrics.}
We evaluate editing quality on MEBench using standard MoLM metrics, grouped by task.
For editing molecule generation, we report \textit{BLEU-4}~\cite{papineni2002bleu} to measure string-level corrections, LEV~\cite{yujian2007normalized} to capture normalized edit distance that calculates the length-normalized Levenshtein distance between two SMILES strings, and MACCS FTS~\cite{zhang2024atomas} to assess Tanimoto substructure similarity via MACCS fingerprint matches. Together, these metrics provide complementary lexical and structural perspectives on the quality of the edits. For editing caption generation, we employ BLEU-2\cite{wieting2019beyond} to measure n-gram overlap, METEOR\cite{banerjee2005meteor} to account for semantic similarity through synonym and stem matching, and ROUGE-1~\cite{lin2004rouge} to evaluate unigram recall, offering a balanced assessment of lexical and semantic fidelity.

\subsubsection{Implementation Details.}
In this section, we will introduce the implementation details of MolEdit. The Multi-Expert Knowledge Adapter's similarity threshold $\epsilon$ is 0.9 for MolFM molecule generation, 0.8 for MoMu molecule generation, and 0.9 for caption generation. The number of experts P is set to 5, with top-$k$ fixed at 1. The distance function $d(\cdot)$ is instantiated as cosine similarity. Following prior work~\cite{chengetal2023edit, wang2024wise}, we edit mid-to-late layers (specifically, layer 4 of the encoder and layer 10 of the decoder in our work) for all tasks and backbones. We edit one piece of knowledge at a time for molecule generation and two pieces for caption generation, due to batch normalization in the MoMu and MolFM molecule encoders. In addition, learning rates are set to 1e-4 for caption generation, 2e-5 for MoMu molecule generation, and 1e-5 for MolFM molecule generation. All experiments were conducted on an NVIDIA A100 server with four GPUs.

\begin{table*}[t!]
    \centering
    \resizebox{0.9\textwidth}{!}{%
    \begin{tabular}{llcccccc}
        \toprule
        \multirow{2}{*}{} & \multirow{2}{*}{} & \multicolumn{3}{c}{\textbf{MoMu}} & \multicolumn{3}{c}{\textbf{MolFM}} \\
        \cmidrule(lr){3-5} \cmidrule(lr){6-8}
         &  & BLEU-2$\uparrow$ & METEOR$\uparrow$ & ROUGE-1$\uparrow$ & BLEU-2$\uparrow$ & METEOR$\uparrow$ & ROUGE-1$\uparrow$\\
        \midrule
        \multirow{6}{*}{\textbf{Rel}} 
        & FT (Encoder) & 0.334~\footnotesize{$(\pm 0.005)$}	& 0.365~\footnotesize{$(\pm 0.006)$}	& 0.472~\footnotesize{$(\pm 0.007)$}
& 0.321~\footnotesize{$(\pm 0.014)$}	& 0.346~\footnotesize{$(\pm 0.019)$}	& 0.448~\footnotesize{$(\pm 0.023)$} \\
        & FT (Decoder) & 0.784~\footnotesize{$(\pm 0.017)$}	& 0.813~\footnotesize{$(\pm 0.014)$}	& 0.854~\footnotesize{$(\pm 0.013)$}
& 0.916~\footnotesize{$(\pm 0.000)$}	& 0.937~\footnotesize{$(\pm 0.000)$}	& 0.958~\footnotesize{$(\pm 0.000)$} \\
        & FT (All) & 0.886~\footnotesize{$(\pm 0.034)$}	& 0.907~\footnotesize{$(\pm 0.030)$}	& 0.925~\footnotesize{$(\pm 0.024)$}
& 0.921~\footnotesize{$(\pm 0.001)$}	& 0.941~\footnotesize{$(\pm 0.001)$}	& 0.960~\footnotesize{$(\pm 0.001)$} \\
        \cmidrule{2-8}
        & MEND & 0.557~\footnotesize{$(\pm 0.034)$}	& 0.569~\footnotesize{$(\pm 0.024)$}	& 0.631~\footnotesize{$(\pm 0.022)$}
& 0.627~\footnotesize{$(\pm 0.025)$}	& 0.652~\footnotesize{$(\pm 0.014)$}	& 0.715~\footnotesize{$(\pm 0.007)$} \\
         & GRACE & \underline{0.928}~\footnotesize{$(\pm 0.000)$}	& \underline{0.947}~\footnotesize{$(\pm 0.000)$}	& \underline{0.965}~\footnotesize{$(\pm 0.000)$}
& \underline{0.928}~\footnotesize{$(\pm 0.000)$}	& \underline{0.947}~\footnotesize{$(\pm 0.000)$}	& \underline{0.965}~\footnotesize{$(\pm 0.000)$} \\
        & \textbf{MolEdit} & \textbf{0.978}~\footnotesize{$(\pm 0.006)$}	& \textbf{0.978}~\footnotesize{$(\pm 0.007)$}	& \textbf{0.982}~\footnotesize{$(\pm 0.005)$}
& \textbf{0.977}~\footnotesize{$(\pm 0.006)$}	& \textbf{0.979}~\footnotesize{$(\pm 0.004)$}	& \textbf{0.983}~\footnotesize{$(\pm 0.004)$} \\
        \midrule
        \multirow{6}{*}{\textbf{Loc}}
        & FT (Encoder) & 0.511~\footnotesize{$(\pm 0.006)$}	& 0.536~\footnotesize{$(\pm 0.010)$}	& 0.604~\footnotesize{$(\pm 0.006)$}
& 0.491~\footnotesize{$(\pm 0.054)$}	& 0.508~\footnotesize{$(\pm 0.065)$}	& 0.587~\footnotesize{$(\pm 0.052)$} \\
        & FT (Decoder) & 0.637~\footnotesize{$(\pm 0.023)$}	& 0.653~\footnotesize{$(\pm 0.030)$}	& 0.699~\footnotesize{$(\pm 0.023)$}
& 0.669~\footnotesize{$(\pm 0.000)$}	& 0.688~\footnotesize{$(\pm 0.000)$}	& 0.732~\footnotesize{$(\pm 0.000)$} \\
        & FT (All) & 0.625~\footnotesize{$(\pm 0.066)$}	& 0.637~\footnotesize{$(\pm 0.063)$}	& 0.688~\footnotesize{$(\pm 0.056)$}
& 0.656~\footnotesize{$(\pm 0.004)$}	& 0.671~\footnotesize{$(\pm 0.006)$}	& 0.721~\footnotesize{$(\pm 0.007)$} \\
        \cmidrule{2-8}
        & MEND & 0.615~\footnotesize{$(\pm 0.010)$}	& 0.633~\footnotesize{$(\pm 0.011)$}	& 0.676~\footnotesize{$(\pm 0.016)$}
& 0.844~\footnotesize{$(\pm 0.005)$}	& 0.856~\footnotesize{$(\pm 0.006)$}	& 0.873~\footnotesize{$(\pm 0.005)$} \\
         & GRACE & \underline{0.875}~\footnotesize{$(\pm 0.013)$}	& \underline{0.883}~\footnotesize{$(\pm 0.006)$}	& \underline{0.904}~\footnotesize{$(\pm 0.005)$}
& \underline{0.893}~\footnotesize{$(\pm 0.002)$}	& \underline{0.899}~\footnotesize{$(\pm 0.001)$}	& \underline{0.917}~\footnotesize{$(\pm 0.002)$} \\
        & \textbf{MolEdit} & \textbf{0.978}~\footnotesize{$(\pm 0.010)$}	& \textbf{0.981}~\footnotesize{$(\pm 0.009)$}	& \textbf{0.983}~\footnotesize{$(\pm 0.008)$}
& \textbf{0.991}~\footnotesize{$(\pm 0.001)$}	& \textbf{0.991}~\footnotesize{$(\pm 0.000)$}	& \textbf{0.993}~\footnotesize{$(\pm 0.000)$} \\
        \bottomrule
    \end{tabular}
    }
    \caption{Main results on MEBench for editing MoMu and MolFM in caption generation, evaluated across Reliability (Rel) and Locality (Loc) dimensions. Each dimension uses three metrics: BLEU-2, METEOR, and ROUGE-1. The best and second-best results are shown in \textbf{bold} and \underline{underlined}, respectively. FT denotes fine-tuning.}
    \label{tab:main_cap}
\end{table*}

\begingroup
\begin{table*}[t!]
    \centering
    \resizebox{0.9\textwidth}{!}{%
    \begin{tabular}{llcccccc}
        \toprule
        \multirow{2}{*}{} & \multirow{2}{*}{} & \multicolumn{3}{c}{\textbf{MoMu}} & \multicolumn{3}{c}{\textbf{MolFM}} \\
        \cmidrule(lr){3-5} \cmidrule(lr){6-8}
         &  & BLEU-2$\uparrow$ & METEOR$\uparrow$ & ROUGE-1$\uparrow$ & BLEU-2$\uparrow$ & METEOR$\uparrow$ & ROUGE-1$\uparrow$\\
        \midrule
        \multirow{5}{*}{\textbf{Rel}} 
        & Moledit        & \textbf{0.978}~\footnotesize{$(\pm 0.006)$} & \textbf{0.978}~\footnotesize{$(\pm 0.007)$} & \textbf{0.982}~\footnotesize{$(\pm 0.005)$}
                         & \textbf{0.977}~\footnotesize{$(\pm 0.006)$} & \textbf{0.979}~\footnotesize{$(\pm 0.004)$} & \textbf{0.983}~\footnotesize{$(\pm 0.004)$} \\
        & w/o EAES       & \underline{0.966}~\footnotesize{$(\pm 0.007)$} & \underline{0.968}~\footnotesize{$(\pm 0.006)$} & \underline{0.974}~\footnotesize{$(\pm 0.006)$}
                         & 0.969~\footnotesize{$(\pm 0.005)$} & 0.970~\footnotesize{$(\pm 0.005)$} & 0.975~\footnotesize{$(\pm 0.005)$} \\
        & w/o MEKA       & 0.773~\footnotesize{$(\pm 0.017)$} & 0.782~\footnotesize{$(\pm 0.015)$} & 0.823~\footnotesize{$(\pm 0.013)$}
                         & 0.803~\footnotesize{$(\pm 0.018)$} & 0.818~\footnotesize{$(\pm 0.017)$} & 0.854~\footnotesize{$(\pm 0.015)$} \\
        & Decoder-only   & 0.955~\footnotesize{$(\pm 0.012)$} & 0.953~\footnotesize{$(\pm 0.011)$} & 0.964~\footnotesize{$(\pm 0.010)$}
                         & \underline{0.972}~\footnotesize{$(\pm 0.004)$} & \underline{0.974}~\footnotesize{$(\pm 0.004)$} & \underline{0.982}~\footnotesize{$(\pm 0.004)$} \\
        & Encoder-only   & 0.404~\footnotesize{$(\pm 0.014)$} & 0.416~\footnotesize{$(\pm 0.015)$} & 0.512~\footnotesize{$(\pm 0.016)$}
                         & 0.413~\footnotesize{$(\pm 0.016)$} & 0.426~\footnotesize{$(\pm 0.017)$} & 0.521~\footnotesize{$(\pm 0.018)$} \\
        \midrule
        \multirow{5}{*}{\textbf{Loc}}
        & Moledit        & \textbf{0.978}~\footnotesize{$(\pm 0.010)$} & \textbf{0.981}~\footnotesize{$(\pm 0.009)$} & \textbf{0.983}~\footnotesize{$(\pm 0.008)$}
                         & \underline{0.991}~\footnotesize{$(\pm 0.001)$} & \underline{0.991}~\footnotesize{$(\pm 0.000)$} & \underline{0.993}~\footnotesize{$(\pm 0.000)$} \\
        & w/o EAES       & 0.924~\footnotesize{$(\pm 0.013)$} & 0.928~\footnotesize{$(\pm 0.012)$} & 0.944~\footnotesize{$(\pm 0.011)$}
                         & 0.973~\footnotesize{$(\pm 0.004)$} & 0.974~\footnotesize{$(\pm 0.004)$} & 0.977~\footnotesize{$(\pm 0.004)$} \\
        & w/o MEKA       & \textbf{0.978}~\footnotesize{$(\pm 0.008)$} & \underline{0.979}~\footnotesize{$(\pm 0.007)$} & \underline{0.981}~\footnotesize{$(\pm 0.007)$}
                         & 0.990~\footnotesize{$(\pm 0.001)$} & 0.988~\footnotesize{$(\pm 0.001)$} & 0.992~\footnotesize{$(\pm 0.001)$} \\
        & Decoder-only   & \underline{0.958}~\footnotesize{$(\pm 0.010)$} & 0.961~\footnotesize{$(\pm 0.010)$} & 0.962~\footnotesize{$(\pm 0.009)$}
                         & 0.977~\footnotesize{$(\pm 0.004)$} & 0.979~\footnotesize{$(\pm 0.004)$} & 0.981~\footnotesize{$(\pm 0.004)$} \\
        & Encoder-only   & 0.947~\footnotesize{$(\pm 0.012)$} & 0.951~\footnotesize{$(\pm 0.011)$} & 0.952~\footnotesize{$(\pm 0.011)$}
                         & \textbf{0.996}~\footnotesize{$(\pm 0.001)$} & \textbf{0.996}~\footnotesize{$(\pm 0.000)$} & \textbf{0.998}~\footnotesize{$(\pm 0.000)$} \\
        \bottomrule
    \end{tabular}
    }
    \caption{Ablation study for editing caption generation under the Reliability (Rel) and Locality (Loc) dimensions. For each dimension, we perform the evaluation by using three metrics: BLEU-2, METEOR, and GOUGE-1. The best and second-best results are shown in bold and underlined, respectively. }
    \label{tab:ab_cap}
\end{table*}

\begin{figure*}[!htbp]
    \centering
    \includegraphics[width=0.9\linewidth]{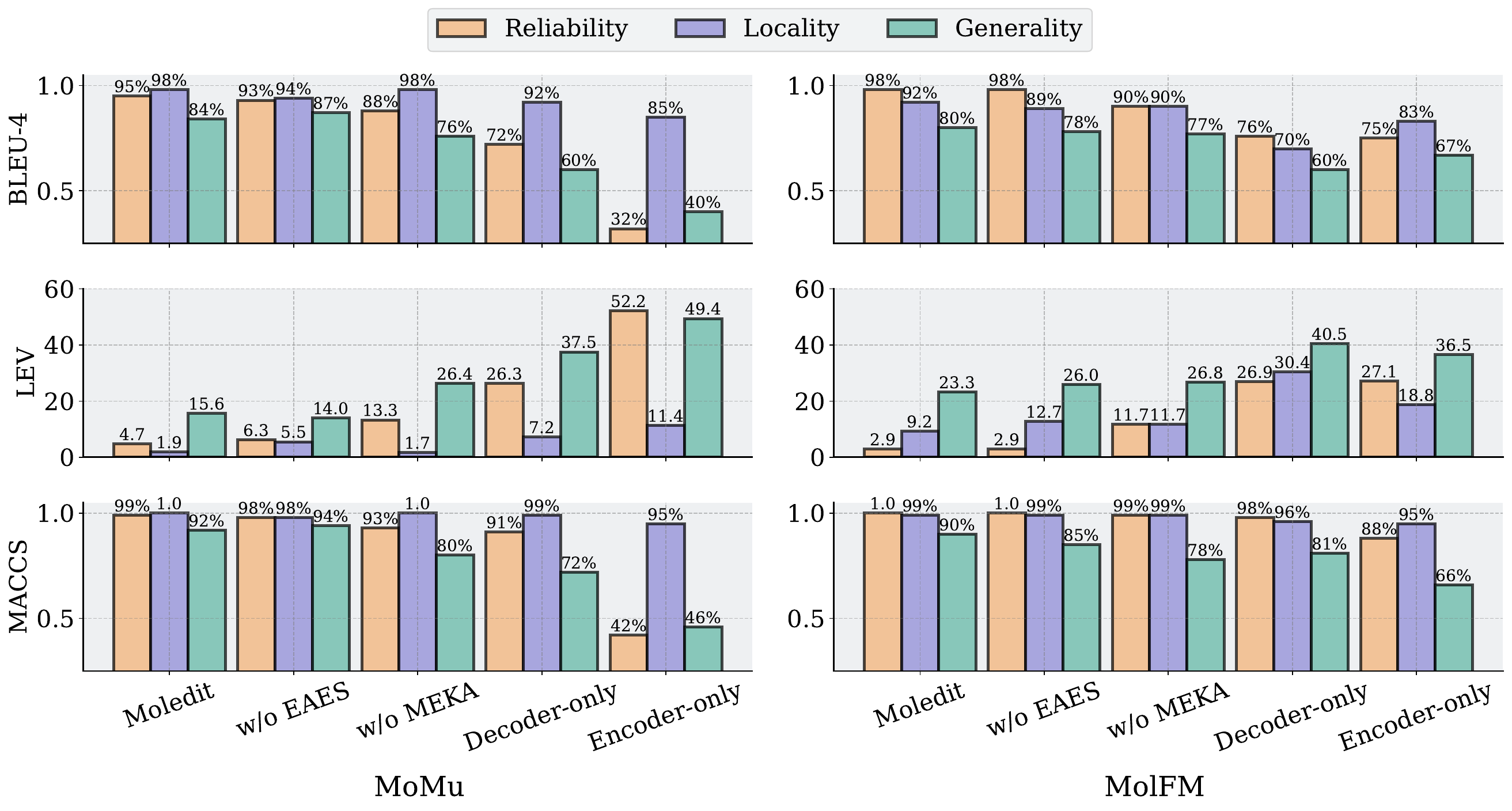}
    \caption{Ablation study for editing molecule generation under the Reliability, Locality, and Generality dimensions. For each dimension, we perform the evaluation by using three metrics: BLEU-4, LEV, and MACCS. EAES denotes Expertise-Aware Editing Switcher while MEKA denotes Multi-Expert Knowledge Adapter.}
    \label{fig:ab_smiles}
\end{figure*}

\begin{figure}[t]
    \centering
    \includegraphics[width=\columnwidth]{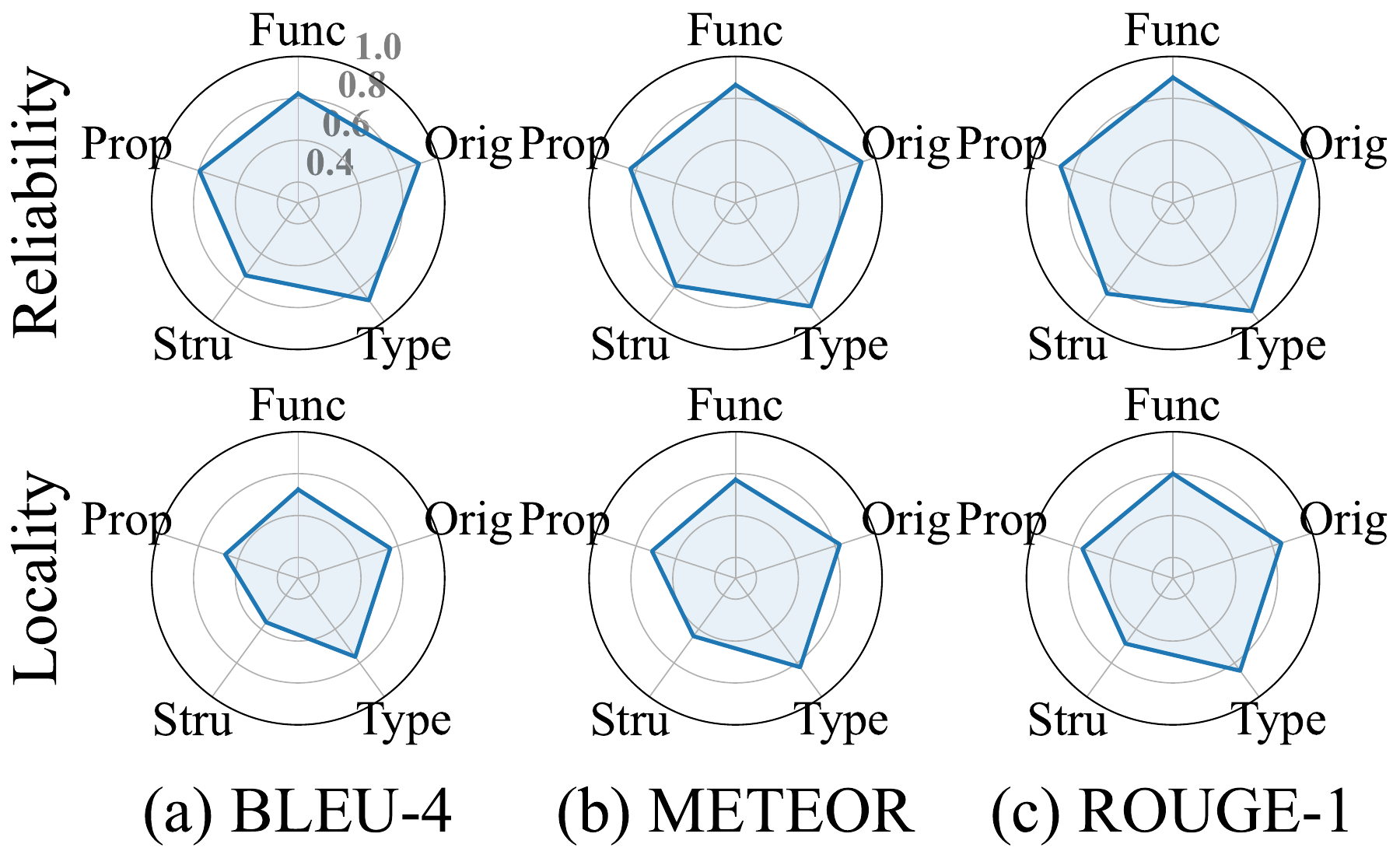}
    \caption{Performance of fine-tuning MoMu on a variation dataset of MEBench. Each subset in this variation dataset targets caption editing where only a single type of expertise requires modification. The expertise is labeled by domain experts and describes different molecular aspects, including (1) \textit{Function} (Func), (2) \textit{Origin} (Orig), (3) \textit{Structure} (Stru), (4) \textit{Type}, and (5) \textit{Property} (Prop).}
    \vspace{-5pt}
    \label{fig:spider}
\end{figure}

\subsection{Main Experiments}
To answer \textbf{RQ1}, we first evaluate MolEdit's performance against all baseline methods on MEBench for both molecule and caption generation tasks.
We make the following observations in Table~\ref{tab:main_smiles} and Table~\ref{tab:main_cap}: (1) For molecule generation editing, MolEdit consistently surpasses all baselines across most metrics. In particular, under the BLEU-4 metric, MolEdit outperforms the second-best approach by up to $18.8\%$ in Reliability, $10.2\%$ in Locality, and $19.1\%$ in Generality. (2) For caption generation editing, MolEdit again achieves superior performance across all metrics. Notably, in terms of BLEU-2, it exceeds the second-best method by up to $5.4\%$ in Reliability and $12.0\%$ in Locality. These results highlight MolEdit’s effectiveness in accurately updating molecule-specific knowledge across different editing tasks.
(3) MolEdit generally performs worse on molecule generation than on caption generation, possibly because editing molecular SMILES strings is inherently more complex and abstract, lacking explicit semantic structure.
(4) MolEdit performs better when editing MoMu compared to MolFM, suggesting that editing becomes more challenging as the underlying model complexity increases.
(5) The performance under the Generality is consistently worse than Reliability, indicating a persistent gap across different representations that encode the same underlying knowledge.
We further observe differences in how baselines behave on MEBench:
(1) Different methods excel in different evaluation dimensions on MEBench. Fine-tuning, while effective at reliable knowledge updates, struggles to preserve irrelevant knowledge, since it updates a large portion of the model's parameters without constraint. MEND performs better at knowledge preservation but underperforms in Reliability, potentially due to the complexity of unstructured molecular knowledge, which poses a challenge for meta-learning approaches. GRACE excels at editing caption generation, but performs poorly in editing molecule generation, likely because it treats captions in a coarse-grained manner rather than capturing fine-grained expertise-level distinctions.
(2) The choice of the edited module significantly influences performance. Fine-tuning the encoder generally enhances performance in editing molecule generation but leads to weaker performance in editing caption generation, suggesting that knowledge is stored in different locations depending on the tasks. Furthermore, the superior performance achieved by editing both modules highlights the presence of synergistic knowledge storage across them.

\begin{figure}[t!]
    \centering
    \includegraphics[width=\columnwidth]{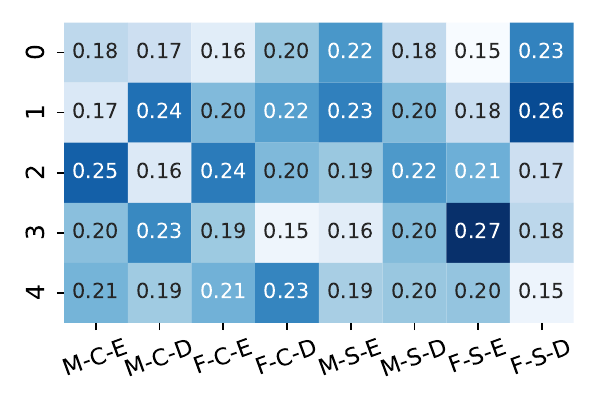}
    \caption{The activation distribution of the experts under different settings. For the label in \textit{x}-axis, "M" denotes MoMu, "F" denotes MolFM; "C" denotes editing caption generation, "S" denotes editing molecule generation; "E" denotes editing encoder, "D" denotes decoder.}
    \label{fig:moe}
\end{figure}

\begin{figure*}[t]
    \centering
    \includegraphics[width=0.9\textwidth]{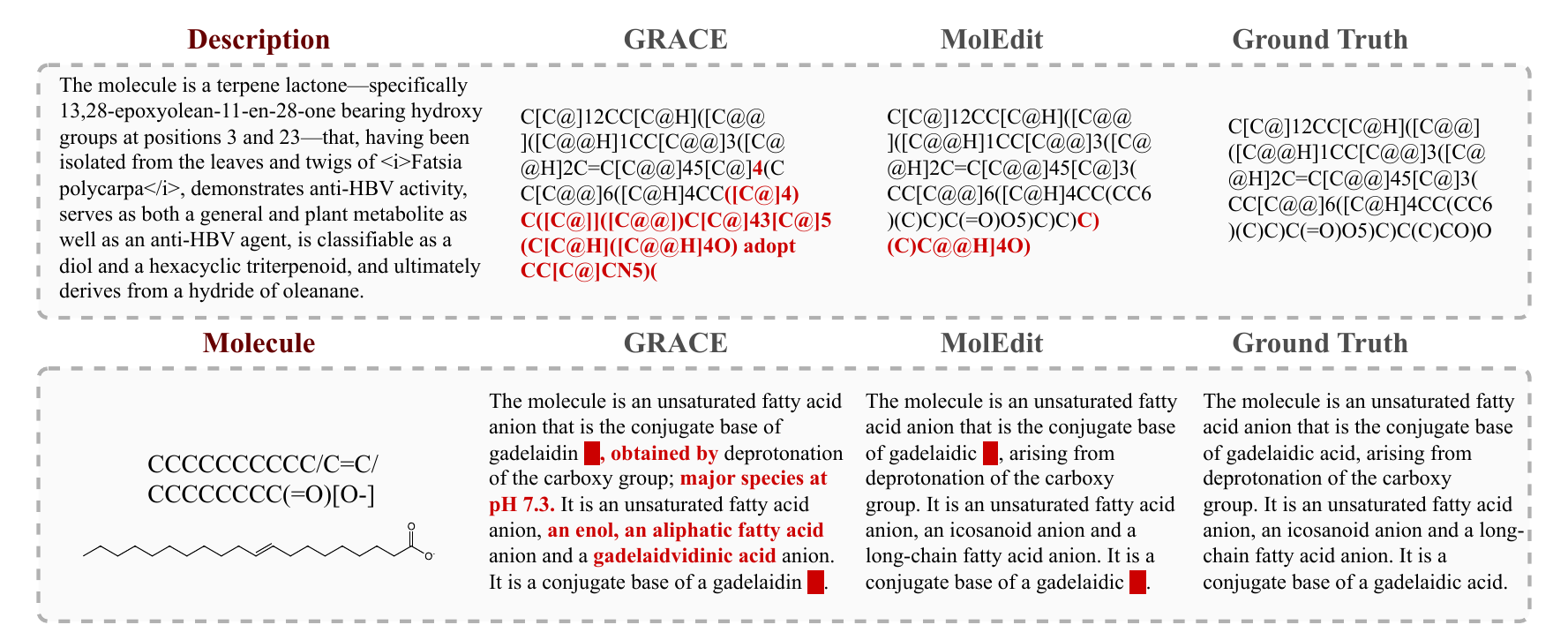}
    \caption{Case study of two tasks on MolFM. Errors are marked in red. GRACE injects invalid SMILES and adds extraneous caption details while MolEdit outputs more valid SMILES and on-target captions. }
    \label{fig:example}
\end{figure*}

\subsection{Ablation Study}
To answer \textbf{RQ2}, we assess the contribution of each module in MolEdit to the performance. We use \textit{w/o MEKA} to denote the use of a plain LoRA adapter without MoE and \textit{w/o EAES} to denote calculating similarity at the sample level rather than the expertise level. Additionally, we evaluate the impact of editing only the encoder or decoder in MolEdit. The results for molecule generation are presented in Figure~\ref{fig:ab_smiles} and the results for caption generation are presented in Table~\ref{tab:ab_cap}. We make the following observations: (1) Both MEKA and EAES contribute to overall performance, validating their effectiveness in addressing molecule-specific challenges in MoLM editing. (2) EAES enhances Locality performance, indicating that it successfully defers untargeted inputs to unedited layers. MEKA improves Reliability performance, suggesting that multiple experts are better equipped to handle multifaceted molecular knowledge. (3) Editing both the encoder and decoder outperforms editing either component individually, highlighting synergistic knowledge storage across these components. (4) Editing the encoder yields markedly lower reliability than editing the decoder, especially for caption generation, suggesting that task-relevant knowledge is concentrated on the decoder side.

\subsection{Rationale Validation}
\label{method_valid}
To answer \textbf{RQ3}, we aim to provide a deeper analysis of each module's rationale. We first validate two interconnected principles behind the multi-expert knowledge adapter: \textbf{(1) The necessity of expertise-wise editing.} We hypothesize that distinct expertise types exhibit heterogeneous sensitivities to editing, creating asymmetric risks of over- and under-editing, thereby necessitating expertise-specific adjustments. To validate this, we curate a specialized dataset from MEBench for molecular caption generation, where each subset focuses on modifying a single targeted expertise, such as Structure set ("two hydroxy groups on the C-30 side-
chain are located at positions 19 and 20.") and Origin set ("It derives from a D-mannitol."). By evaluating fine-tuning performance on this dataset (Figure~\ref{fig:spider}), we observe that editing sensitivities vary across expertise domains, with structure edits being the most challenging. This divergence underscores the importance of adopting expertise-wise editing strategies.
%
\textbf{(2) The effectiveness of expertise-wise editing.} We then examine whether MEKA actually learns to separate and route expertise during training by analyzing the expert activation distribution of the MoE, as shown in Figure~\ref{fig:moe}. The results indicate that all experts consistently exhibit high activation rates, demonstrating that the MoE effectively isolates and routes different expertise during editing rather than collapsing onto a single route. 

Next, we validate \textbf{the effectiveness of expertise-aware editing switcher (EAES)} in selectively activating relevant knowledge while suppressing untargeted information. As shown in Table~\ref{tab:switch}, EAES achieves significantly higher switching accuracy than non-expertise-aware alternatives, with improvements of up to 90.7\%. 
By validating these principles, we demonstrate that MolEdit’s modular design effectively addresses molecule-specific editing challenges, resulting in enhanced performance.

\subsection{Case Study}
We present case studies for two tasks using MolFM, comparing the outputs of GRACE and MolEdit on two representative examples in Fig~\ref{fig:example}. For molecule generation, GRACE frequently introduces extraneous or invalid tokens into the SMILES strings, such as incorrect brackets and the non-SMILES token “adopt” (highlighted in red). For caption generation, GRACE also hallucinates irrelevant caption details (e.g., “major species at pH 7.3” and “enol”), likely remnants from incomplete deletion of prior descriptions. In contrast, MolEdit produces outputs that closely align with the ground truth with minimal spillover, highlighting its robustness and precision in knowledge editing.

\begin{table}[t!]
\centering
\resizebox{\columnwidth}{!}{%
\begin{tabular}{@{} l *{4}{c} @{}} 
\toprule
& M-Cap & F-Cap & M-Mol & F-Mol \\
\midrule
\textbf{Accuracy} & 0.827 & 0.772 & 0.635 & 0.529 \\
{$\mathbf{\Delta\uparrow}$} & 65.5\% & 54.4\% & 90.7\% & 58.9\% \\
\bottomrule
\end{tabular}
}
\caption{Accuracy of Expertise-Aware Editing Switcher under different settings. $\mathbf{\Delta\uparrow}$ represents the improvement in accuracy compared to switchers that do not consider expertise. For abbreviations, "M" denotes MoMu, "F" denotes MolFM; "Cap" denotes editing caption generation, "Mol" denotes editing molecule generation.}
\label{tab:switch}
\end{table}

\balance
\section{Conclusion}
In this paper, we make the first attempt to edit MoLMs for both molecule and caption generation. To address the unique challenges associated with molecular editing, we propose MolEdit, a novel framework which consists of two key components: (1) MEKA, which directs molecular knowledge to specialized editing experts, enabling fine-grained control over multi-faceted updates; and (2) EAES, which maintains a memory bank of edited molecular expertise to activate MEKA only for highly relevant inputs. To enable systematic evaluation, we introduce MEBench, a comprehensive benchmark that assesses three critical dimensions: Reliability, Locality, and Generality. Extensive experiments demonstrate the effectiveness of MolEdit, prompting future work on editing MoLMs.   

\section*{Acknowledgment}
This work is supported in part by the National Science Foundation (NSF) under grants IIS-2144209, IIS-2223769, CNS-2154962, BCS2228534, and CMMI-2411248; the Office of Naval Research (ONR) under grant N000142412636; and the Commonwealth Cyber Initiative (CCI) under grant VV-1Q24-011, and the gift funding from Netflix and Snap.


\newpage
\section*{Ethical Considerations}
This work focuses on methodology for correcting errors in knowledge stored in molecule language models using public datasets and benchmarks. While we see limited ethical risks, potential concerns include misuse of editing to amplify unsafe molecular knowledge and inadvertent propagation of dataset biases. To mitigate these potential problems, we restrict evaluations to benign data, where the curated benchmark only provide benign targets. Overall, we view the ethical risks as minimal and manageable under standard responsible-use practices.

\bibliographystyle{ACM-Reference-Format}
\bibliography{sample-base}


\end{document}